\documentclass{article}

\usepackage{microtype}
\usepackage{graphicx}
\usepackage{subfigure}
\usepackage{booktabs} 
\usepackage{xcolor} 
\usepackage{import}
\usepackage{chapterbib}

\usepackage{hyperref}

\usepackage[accepted]{icml2020}

\usepackage{pgf}
\usepackage{tikz}
\usepackage{collcell}

\usepackage{xspace}
\newcommand*{\eg}{\textit{e.g.}\@\xspace}
\newcommand*{\ie}{\textit{i.e.}\@\xspace}

\newcommand*{\MinNumber}{0.16}%
\newcommand*{\MidNumber}{0.23} %
\newcommand*{\MaxNumber}{.830}%
\def\OmitZero#1.#2.#3!{%
  \ifx\relax#2\relax     
    #1%
  \else                  
    \ifnum#1=0           
      .#2%
    \else                
      #1.#2%
    \fi
  \fi}

\definecolor{blue1}{cmyk}{.1284,.1468,0,0.145}
\definecolor{red1}{cmyk}{0,.49,.5458,0.0157}

\newcommand{\ApplyGradient}[1]{%
    \ifdim #1 pt > \MidNumber pt
    \pgfmathsetmacro{\PercentColor}{max(min(100.0*(#1 - 
    \MidNumber)/(\MaxNumber-\MidNumber),100.0),0.00)} %
    \hspace{-0.33em}\colorbox{red1!\PercentColor!blue1}{#1}
    \else
    \pgfmathsetmacro{\PercentColor}{max(min(100.0*(\MidNumber - 
    #1)/(\MidNumber-\MinNumber),100.0),0.00)} %
    \hspace{-0.33em}\colorbox{white!\PercentColor!blue1}{#1}
    \fi
}
\newcolumntype{G}{>{\collectcell\ApplyGradient}c<{\endcollectcell}}

\usepackage{adjustbox}
\usepackage{array}

\newcolumntype{R}[2]{%
    >{\adjustbox{angle=#1,lap=\width-(#2)}\bgroup}%
    l%
    <{\egroup}%
}
\newcommand*\rotz{\multicolumn{1}{R{0}{-1em}}}
\newcommand*\rot{\rotatebox{45}}

\begin{document}

\icmltitlerunning{Adding Seemingly Uninformative Labels Helps in Low Data Regimes}

\twocolumn[
\icmltitle{Adding Seemingly Uninformative Labels Helps in Low Data Regimes}

\icmlsetsymbol{equal}{*}

\begin{icmlauthorlist}
\icmlauthor{Christos Matsoukas}{KTH,SciLifeLab,AZ}
\icmlauthor{Albert Bou I Hernandez}{KTH}
\icmlauthor{Yue Liu}{KTH,SciLifeLab}
\icmlauthor{Karin Dembrower}{KI,GH}
\icmlauthor{Gisele Miranda}{KTH,SciLifeLab}
\icmlauthor{Emir Konuk}{KTH,SciLifeLab}
\icmlauthor{Johan Fredin Haslum}{KTH,SciLifeLab,AZ}
\icmlauthor{Athanasios Zouzos}{KS}
\icmlauthor{Peter Lindholm}{KI}
\icmlauthor{Fredrik Strand}{KI,KS}
\icmlauthor{Kevin Smith}{KTH,SciLifeLab}
\end{icmlauthorlist}

\icmlaffiliation{KTH}{School of Electrical Engineering and
Computer Science, KTH Royal Institute of Technology, Stockholm, Sweden}
\icmlaffiliation{SciLifeLab}{Science for Life Laboratory, Stockholm, Sweden}
\icmlaffiliation{AZ}{AstraZeneca, Gothenburg, Sweden}
\icmlaffiliation{KI}{Karolinska Institutet, Stockholm, Sweden}
\icmlaffiliation{KS}{Karolinska University Hospital, Stockholm, Sweden}
\icmlaffiliation{GH}{Capio Sankt Göran Hospital, Stockholm, Sweden}

\icmlcorrespondingauthor{Christos Matsoukas}{matsou@kth.se}

\icmlkeywords{machine learning, deep learning, ICML, segmentation, low data regimes, limited data, limited labels, complementary labels, medical, medical imaging, medical data, tumor segmentation, SAW, SAW-S}

\vskip 0.3in
]

\printAffiliationsAndNotice{}  

\begin{abstract}

    Evidence suggests that networks trained on large datasets generalize well not solely because of the numerous training examples, but also class diversity which encourages learning of enriched features.
    This raises the question of whether this remains true when data is scarce \textit{-- is there an advantage to learning with additional labels in low-data regimes?}
    In this work, we consider a task that requires difficult-to-obtain expert annotations: tumor segmentation in mammography images.
    We show that, in low-data settings, performance can be improved by complementing the expert annotations with seemingly uninformative labels from non-expert annotators, turning the task into a multi-class problem.
    We reveal that these gains increase when less expert data is available, and uncover several interesting properties through further studies. 
    We demonstrate our findings on \textsc{Csaw-S}, a new dataset that we introduce here, and confirm them on two public datasets.
    
\end{abstract}

\section{Introduction}
\label{intro}

\begin{figure}[ht]
\begin{center}
\centerline{\includegraphics[width=\columnwidth]{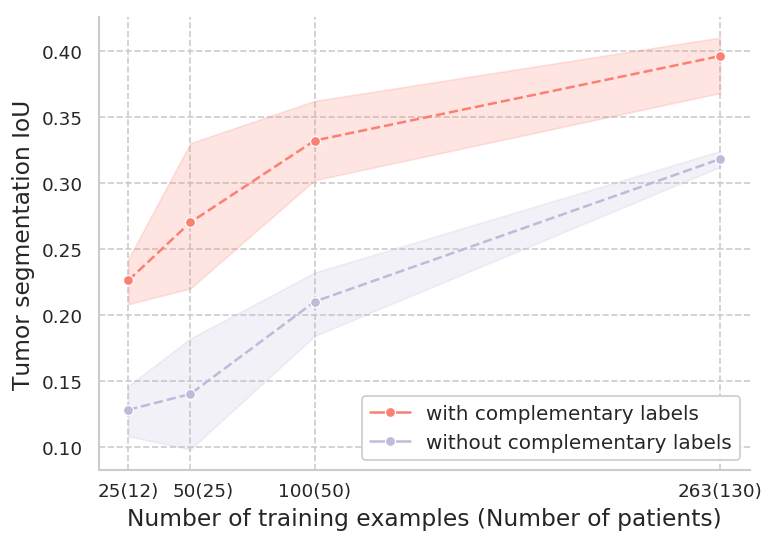}}
\vspace{-4mm}
\caption{We consider low-data regime tasks where expert data is difficult to obtain, such as tumor segmentation in mammography images from the \textsc{Csaw-S} dataset. A network trained with only expert annotations of the tumors (blue) is significantly outperformed by the same network which is given additional complementary non-expert annotations (red) of breast anatomy (\textit{e.g.}~skin, pectoral muscle, nipple), turning the task into a multi-class problem.
Performance is measured for varying numbers of training images/patients by intersection over union (IoU) over five repetitions (shaded area indicates 95\% CI).}
\label{fig:ANON_IoU_absolut}
\end{center}
\end{figure}

When abundant training data is available, deep learning approaches have achieved remarkable feats, especially in areas like computer vision and natural language processing. 
However, even within these well-studied domains, there exist many important problems for which data is scarce. 

In medicine, large well-labeled datasets are difficult to come by for a variety of reasons. 
Often, legal regulations and privacy concerns prohibit the distribution of data. Or there simply may not be enough patients to collect the data from. This may be due to the rare nature of a disease that only affects a limited population, or due to the logistics of collecting data from different administrative regions.
Sometimes the techniques used for diagnosis are expensive or require painful procedures (\textit{e.g.}~biopsies), and thus data is only collected when deemed absolutely necessary.
Finally, there are cases where sufficient data may be available, but crucial expert-level annotations are prohibitively expensive or difficult to obtain.
This problem is common in other domains such as astronomy and biology where either data or expert knowledge is difficult to obtain.

In this work, \textit{we observe that 
performance on expert image segmentation tasks can be significantly improved by adding seemingly uninformative annotations during training}. Although we cannot offer a definitive explanation for this effect, we surmise that the class diversity encourages learning of enriched features that capture complementary information to the main task. The additional labels do not contain any direct evidence for the expert task, but they may give indirect support by providing revealing information about other content in the image.

The principal benefit of this approach is that model performance can be increased without changing the architecture or collecting additional expert training examples. 
The additional labels can be obtained inexpensively and do not need to be of high quality.
This allows one to achieve better performance when it is simply not possible to obtain more data, and to optimize the costs of the data acquisition process by considering the trade-offs between collecting new examples, expert annotations, and non-expert annotations.

Many researchers share an intuition that adding information to the learning targets improves performance, and in fact many related techniques discussed in this work are based on this principle (\eg distillation, multi-task learning). 
Our contribution is to demonstrate and gain insights into this principal for complementary labels, which is not yet well understood.

As our main case study, we introduce \textsc{Csaw-S}, a dataset of mammography images which includes expert annotations of tumors and non-expert annotations of breast anatomy and artifacts in the image  (described in Section \ref{applications}).
Using this dataset we experimentally show that tumor segmentation performance, a task that requires expert annotations, is significantly improved by providing additional complementary non-expert labels (\textit{e.g.}~skin, nipple, pectoral muscle). We further show that this benefit becomes more prominent as data becomes more scarce.

We validate our findings by demonstrating that the observed effect holds in other domains, using public datasets including \textsc{Cityscapes}
and \textsc{Pascal Voc} in Section \ref{experiments}. 
Furthermore, we performed a number of additional studies to gain further insights into the effects of complementary labels, such as the dependence on the number of labels and the relative importance of label types. Our contributions are summarised as follows:
\begin{itemize}
    \vspace{-3mm}
    \item We show empirical evidence that \textit{inexpensive complementary labels improve model performance in low-data regimes}. 
    \vspace{-1mm}
    \item We observe this effect in \textit{1)} a high-impact medical task where training examples are difficult to acquire and expert annotations are expensive and, \textit{2)} two well-studied public datasets.
    \vspace{-1mm}
    \item We conduct a series of  studies that reveal further insights about this phenomenon. We show \textit{a)} how the effect lessens as data increases \textit{b)} that complementary labels provide robustness to annotator bias \textit{c)} the effectiveness of different labels \textit{d)} trivial labels are not useful \textit{e)} performance increases with more labels \textit{f)} low-quality labels are nearly as good as high-quality labels \textit{g)} complementary labels increase training stability \textit{f)} complementary labels provide some robustness to domain shifts.
    \vspace{-1mm}
    \item We release the \textsc{Csaw-S} dataset used in this study to the public, which contains valuable mammography images with labels from multiple experts and non-experts that can be used to replicate our study and for other segmentation tasks.
    \vspace{-3mm}
\end{itemize}

Finally, to promote transparency and reproducibility, we share our open-source code, available at
\href{https://github.com/ChrisMats/seemingly_uninformative_labels}{github.com/ChrisMats/seemingly\_uninformative\_labels} and \textsc{Csaw-S} at \href{https://github.com/ChrisMats/CSAW-S}{https://github.com/ChrisMats/CSAW-S}.

\section{Related Work}
\label{related}

Perhaps the most well-established method of dealing with insufficient training data is to learn transferable representations in a similar domain where data is more abundant, a technique routinely executed by means of pretraining on \textsc{ImageNet}. The underlying assumption for this approach is that the domain gap is small, \ie, the distribution the model is pretrained on has structural similarities to the target task's conditional probability distribution \cite{representations_bengio}. Unfortunately, this assumption does not necessarily hold for all tasks \cite{representation_genercit_to_spec}, for example between natural and medical images. 
\citeauthor{transfusion} have shown that \textsc{ImageNet} pretraining offers only marginal improvement for some medical tasks, mainly attributed to better weight scaling and initialization \cite{transfusion}. 

\citeauthor{rethinking_imagenet_pretrain} showed that
initialization with \textsc{ImageNet} yields no gains compared to random initialization in the big-data regime \cite{rethinking_imagenet_pretrain}. 
Interestingly, as they moved towards the low-data regime they noticed benefits from \textsc{ImageNet} pretraining began to appear for \textsc{Pascal Voc} but not for \textsc{Ms-Coco}. The authors link this effect to the relatively low number of classes and object instances in the \textsc{Pascal Voc} dataset. In other words, \textsc{ImageNet} pretraining helps more when the downstream task is less diverse. \citeauthor{representation_genercit_to_spec}~argue that \textsc{ImageNet} pretrained models benefit more from the diversity of classes than the number of training examples. 
Finally, it has been established that label correlation can increase accuracy in multi-label classification tasks by providing information regarding the interactions among them \cite{label_correlation}.
In line with these insights, we conjecture that more diverse annotations yield better representations and consequently better performance in scarce data settings.

Beyond transfer learning, modern methods such as unsupervised or weakly/semi-supervised approaches, can learn representations yielding comparable performance to supervised learning \cite{jing2019self} 
provided the domain gap is relatively small. 
Recently, \citeauthor{he2019momentum} reported downstream task performances even better than supervised pretraining \cite{he2019momentum}. Unfortunately, as in the supervised case, these methods require large amounts of data, which is a challenge in medical tasks \cite{Hesamian2019}. 

Data augmentation techniques are widely employed when training models with limited data. In their seminal work, \citeauthor{unet} have shown the effectiveness of suitable augmentations for medical image segmentation. Learning augmentations during training \cite{autoaugment, zhao2019data} and GAN-based augmentation \cite{gan_based_augm, mondal2018few} can also be employed in order to alleviate the effects of data scarcity.

Promising results have also been shown by $k$-shot methods \cite{roy2020squeeze} in low data regimes where certain classes only have a few or no representation in the training set. However, the domain gap and data scarcity are also significant problems in the $k$-shot settings as most of the methods in the literature rely on \textsc{ImageNet} pretraining, which limits it's applicability in non-natural image domains.

In this work, we argue that adding new labels that complement the ones provided for the principal task improves generalization in low-data regimes.
Side information is a term for extraneous information, often a different modality than the source data, that can be exploited for a principal task \cite{side_info_convolution}.
For example, \cite{Tian_2015_CVPR} used additional datasets to introduce side information --expressed as scene and pedestrian attributes-- to pedestrian classification.

\section{Complementary Labels}
\label{methods}

Our central idea is simple but effective in practice. We argue that seemingly uninformative complementary labels, used as additional learning targets, have a direct impact on the model's generalization for image segmentation in low data regimes.

\begin{figure}[t]
\begin{center}
\begin{tabular}{@{}c@{}c@{\hspace{1mm}}c@{}}
 \includegraphics[height=4.7cm]{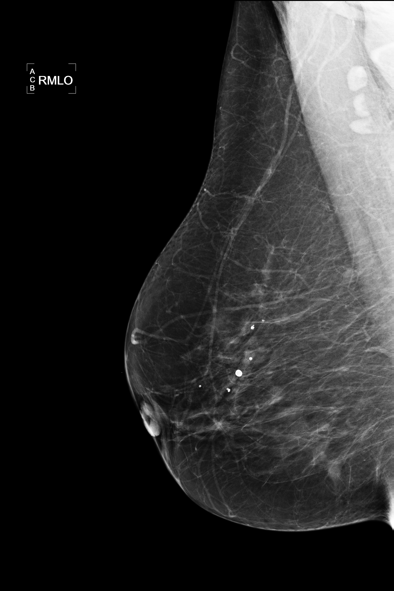}    &
 \includegraphics[height=4.7cm]{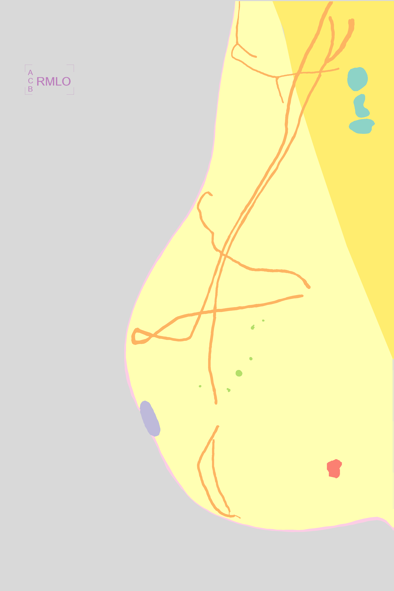}    &
\includegraphics[height=4.7cm]{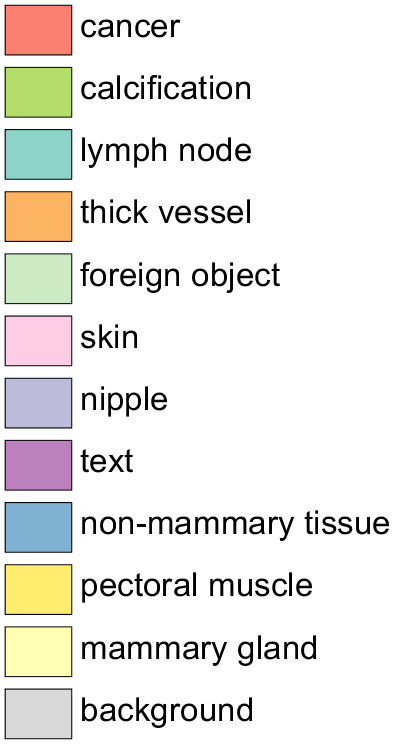}\\
\end{tabular}
\vspace{-3mm}
\caption{The \textsc{Csaw-S} dataset, released here, contains 342 mammograms with expert radiologist labels for cancer and \textit{complementary labels} of breast anatomy made by non-experts. The non-expert labels are imperfect and in some cases may seem uninformative, yet they provide useful cues for segmentation of the tumor.}
\label{fig:complementary_labels}
\end{center}
\vskip 2mm
\end{figure}

These new annotations are often \textit{seemingly uninformative} in the sense that they do not contain any direct information about the object of interest.
However, they do contain useful information describing other semantically meaningful objects present in the image. These labels complement the expert labels by providing rich contextual information, so we refer to them as \textit{complementary labels}. For the task of locating tumors in mammograms, complementary labels might include the \textit{skin}, \textit{pectoral muscle}, and other parts of anatomy  (Figure \ref{fig:complementary_labels}). Complementary labels do not require expert knowledge and do not need to be particularly accurate, so crowdsourcing or other low-cost solutions can be employed to collect them.

The advantage of complementary labels is that they are easy to obtain and the performance boost comes without major changes to the model architecture or additional training examples. On the other hand, as more expert data are available, complementary labels yield diminishing returns (Figure~\ref{fig:ANON_IoU_absolut}). So consideration must be given to the performance-cost trade-off between obtaining inexpensive labels for existing data or collecting new data with expert labels.

Our observation invites an obvious question -- \textit{why} does adding complementary annotations lead to better performance? Although we cannot provide a definitive answer, we offer two plausible explanations:
\vspace{-3mm}
\begin{enumerate}
    \item \textit{Complementary labels encourage learning of enriched representations.}
    With sufficient training examples, deep neural networks learn rich features that represent not only the object of interest, but also model the diversity of shapes and textures in the background. In classification and detection tasks, evidence has shown that networks learn to exploit information from the background \cite{wolf_dogs_explainable,concept_expanations}.
    But in the low data regime, the network struggles to model the background because the data is insufficient to capture such a diverse distribution. Complementary labels help the network make sense of reduced background data by structuring it into more meaningful sub-classes with less individual variation.
    Analogous explanations have been hypothesized in \cite{transfer_imagenet_efros,representation_genercit_to_spec} for classification and object detection \cite{rethinking_imagenet_pretrain}, where it is argued that providing fewer labels per image has a similar effect to removing training examples.
    
    \item \textit{Complementary labels help to model interactions between objects.}
    Objects that are near, interact with, or look similar to the target object contain information regarding the correlation and interaction that the target has with its environment. Providing complementary labels allows the network to exploit interactions and correlations between these additional labels and the target, resulting in better generalization. A similar line of thinking explains how knowledge distillation benefits from interactions between labels through soft labeling \cite{distillation,borh_again_networks}. Likewise, in \cite{label_correlation} correlations between rare labels are exploited to achieve better performance in a multi-label prediction task.  
\end{enumerate}

\vspace{-1mm}
In the following sections, we provide experimental evidence supporting the benefits of complementary labels in low data regimes and explore practical questions surrounding their use and effects on generalization. In particular, we address:

\vspace{-1mm}
\begin{itemize}
    \item What is the effect of changing the number of training examples/expert annotations?
    \vspace{-1mm}
    \item What constitutes a good complementary label?
    \vspace{-1mm}
    \item What is the effect of adding new complementary labels?
    \vspace{-1mm}
    \item How does the quality of complementary labels affect performance?
    \vspace{-1mm}
    \item Do complementary labels provide robustness: \textit{1)} to annotator bias, \textit{2)} to domain shifts in the training data,  and \textit{3)} training stability?
\end{itemize}
\vspace{-1mm}

Through the following experiments on \textsc{Csaw-S} and two well-known public datasets, we attempt to characterize these properties of complementary labels.

\section{Experiments on Medical Images}
\label{applications}

In this section we investigate how the addition of complementary labels affects the model's performance in a low data setting on an expert medical task.

\begin{figure}[t!h]
\begin{center}
\begin{tabular}{@{}c@{\hspace{0.5mm}}c@{\hspace{0.5mm}}c@{\hspace{0.5mm}}c@{}}
\includegraphics[width=20mm]{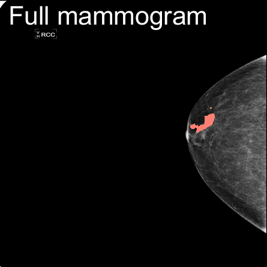} &
\includegraphics[width=20mm]{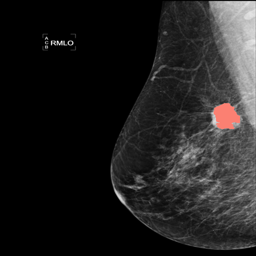} &
\includegraphics[width=20mm]{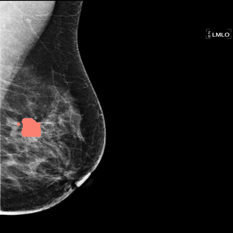} &
\includegraphics[width=20mm]{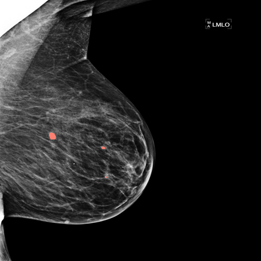} \\[.5mm]
\includegraphics[width=20mm]{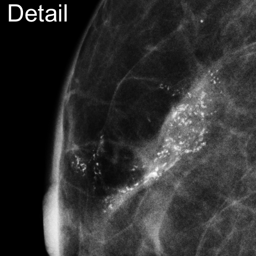} &
\includegraphics[width=20mm]{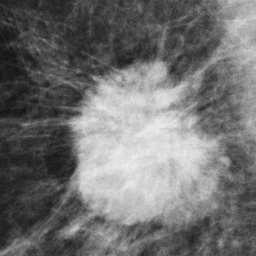} &
\includegraphics[width=20mm]{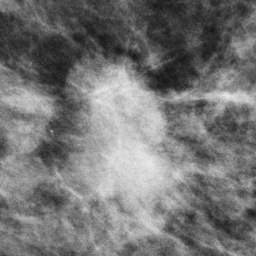} &
\includegraphics[width=20mm]{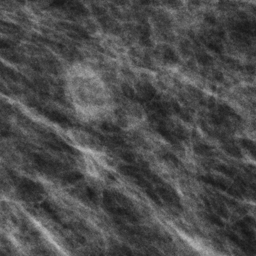} \\[.5mm]
\includegraphics[width=20mm]{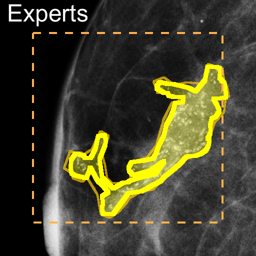} &
\includegraphics[width=20mm]{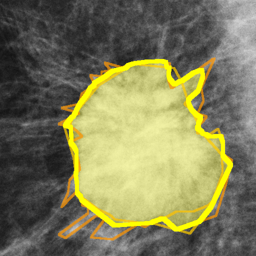} &
\includegraphics[width=20mm]{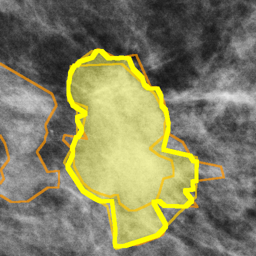} &
\includegraphics[width=20mm]{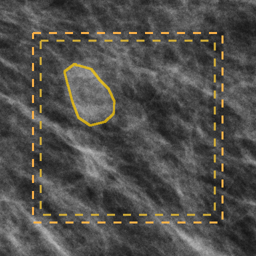} \\[.5mm]
\includegraphics[width=20mm]{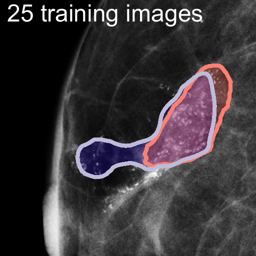} &
\includegraphics[width=20mm]{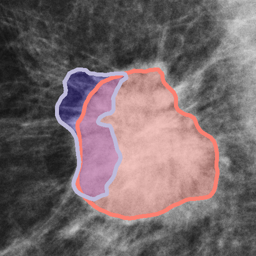} &
\includegraphics[width=20mm]{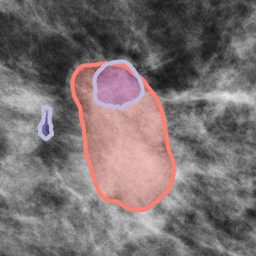} &
\includegraphics[width=20mm]{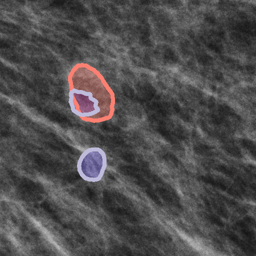} \\[.5mm]
\includegraphics[width=20mm]{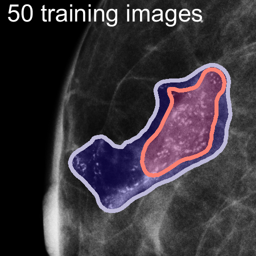} &
\includegraphics[width=20mm]{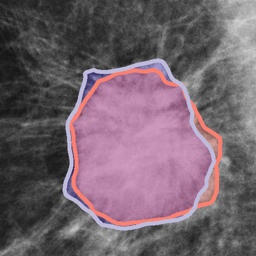} &
\includegraphics[width=20mm]{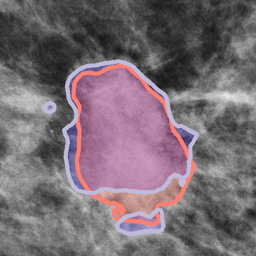} & 
\includegraphics[width=20mm]{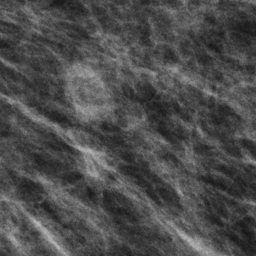} \\[.5mm]
\includegraphics[width=20mm]{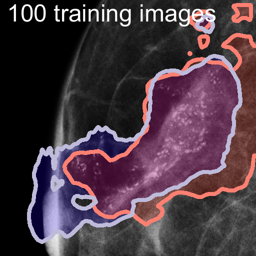} &
\includegraphics[width=20mm]{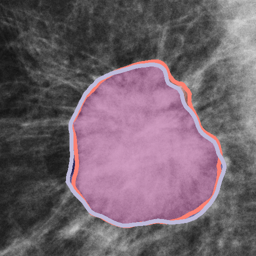} &
\includegraphics[width=20mm]{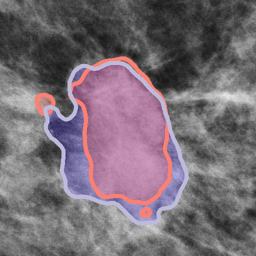} &
\includegraphics[width=20mm]{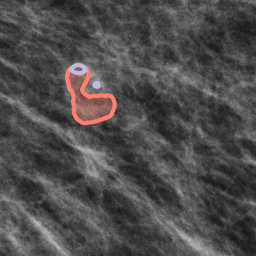} \\[.5mm]
\includegraphics[width=20mm]{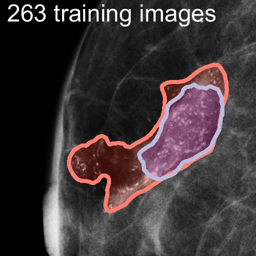} &
\includegraphics[width=20mm]{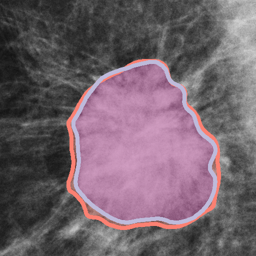} &
\includegraphics[width=20mm]{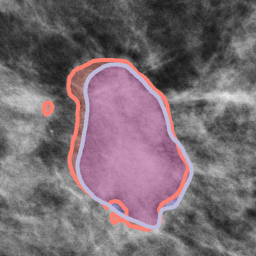} &
\includegraphics[width=20mm]{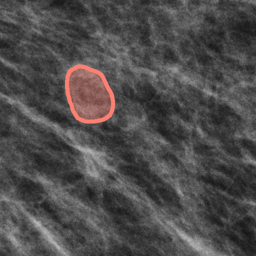} \\[1mm]
\end{tabular}
\includegraphics[width=\columnwidth]{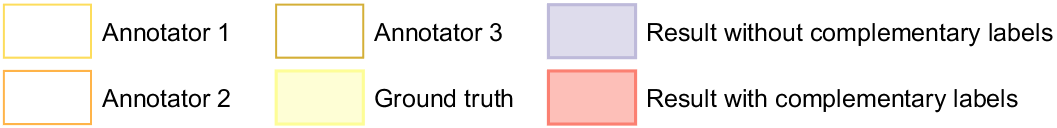}
\vspace{-5mm}
\caption{Segmentation results on \textsc{Csaw-S} show complementary labels improve performance in low-data settings. From top to bottom: the full mammogram, a detailed image of the tumor region, the expert annotations and ground truth (\textit{yellow}), predictions from networks trained with only the tumor labels (\textit{blue}) and predictions from a network trained with both tumor labels and complementary labels (\textit{red}). Experiments are repeated using small training sets of varying size, $N=\{$25, 50, 100, 263$\}$, the best result from 5 runs is shown for each model. Expert annotations appear in shades of gold; a dashed box indicates an annotator did not find a tumor. Ground truth is determined where at least two annotators agree cancer is present (consensus in column 4 is \textit{no tumor}).}
\label{fig:ANON_viualizations}
\end{center}
\vspace{2mm}
\end{figure}

\subsection{The \textsc{Csaw-S} Dataset}
\label{ANON-S}

The \textsc{Csaw-S} dataset is a companion subset of \textsc{Csaw}, a large cohort of  mammography data gathered from the entire population of Stockholm invited for screening between 2008 and 2015, which is available for research \cite{dembrower2019multi}.
We release the \textsc{Csaw-S} subset containing mammography screenings from 172 different patients with annotations for semantic segmentation. The patients are split into a test set of 26 images from 23 patients and training/validation set containing 312 images from 150 patients. Further details regarding the collection and pre-processing of the data can be found in the Appendix.

The training/validation images are accompanied by cancer annotations by an expert radiologist, \textsc{expert 1}, and the test images come with cancer annotations from two additional radiologists, \textsc{expert 2} and \textsc{expert 3}. 
Complementary labels are provided for the entire dataset in the form of full pixel-wise masks of each image corresponding to 11 additional highly imbalanced classes representing breast anatomy and other objects (see Figure \ref{fig:complementary_labels} and Appendix for details). \textit{The complementary annotations were sourced from non-experts with no medical training, and therefore contain errors}. Complementary annotations from \textsc{expert 1} and \textsc{expert 2} are also provided for the test set. The ground truth for the expert task is determined as regions where at least two experts agree cancer is present (Fig.~\ref{fig:ANON_viualizations}).

\begin{table}[t]
\vspace{-2mm}
        \caption{Expert and model agreement on \textsc{Csaw-S} (IoU)} 
        \label{tab:confusion}
        \centering
        \resizebox{.65\columnwidth}{!}{ 
            \begin{tabular}{r*{6}{G}}
                \rotz{} &
                \rotz{\rot{\textsc{Expert 1}}} & 
                \rotz{\rot{\textsc{Expert 2}}} & 
                \rotz{\rot{\textsc{Expert 3}}} &
                \rotz{\rot{with comp.$^*$}} & 
                \rotz{\rot{without comp.$^*$}} \smallskip \\

                Ground Truth \hspace{2mm}      & 0.69 & 0.66 & 0.83 & 0.33 & 0.21     \\ 

                \textsc{Expert 1}  \hspace{2mm}    & \rotz{} & 0.67 & 0.68 & 0.31 & 0.21     \\ 

               \textsc{Expert 2} \hspace{2mm}   &  \rotz{} & \rotz{} & 0.66 & 0.29 & 0.17     \\ 

                \textsc{Expert 3} \hspace{2mm}    & \rotz{} & \rotz{} & \rotz{} & 0.34 & 0.21     \\                 
            \end{tabular}
        }
        \newline
                    \vspace{1mm}    {\tiny $^*$Mean IoU for models trained with 50 patients.}
        \vspace{3mm}
    \end{table}

\subsection{Experimental Setup}
We split the train/validation sets by patient, 130/20. This  resulted in 263/49 images per set.
Many of our experiments investigate how performance depends on the size of the training set. For these cases, we create subsets of $N=\{$25, 50, 100, 263$\}$ images by sampling the training set at the patient level. We perform five runs for each experiment by sampling with replacement 5 times, and report the average performance along with the 95\% confidence interval.

Our goal is to improve expert task performance by including complementary non-expert labels. 
Hence, we define two training settings to test this
\begin{enumerate}
    \vspace{-3mm}
    \item \textit{with complementary labels} -- where expert labels and complementary labels are provided to train the network
    \vspace{-5mm}
    \item \textit{without complementary labels} -- where only expert labels are provided.
    \vspace{-3mm}
\end{enumerate}

These cases are compared throughout Sections \ref{applications} and \ref{experiments}.

\subsection{Implementation Details}

We use DeepLab3 \cite{deeplabv3} with ResNet50 \cite{resnet} as the  backbone for all experiments. 
Following \citeauthor{rethinking_imagenet_pretrain} and \citeauthor{transfusion}, we initialize all models with \textsc{ImageNet} pretrained weights and we replace BatchNorm layers  with GroupNorm layers \cite{groupnorm}. 
We use an ADAM \cite{adam} optimizer throughout our experiments.

Due to memory limitations and the high resolution of mammograms, we train using $512\times512$ patches. To ensure good representation in the training data, for every full image we sample a center-cropped patch from 10 random locations belonging to each of the 12 classes (the same for training with and without complementary labels).

To alleviate overfitting issues associated with extreme low data regimes, we employ an extensive set of augmentations including rotations and elastic transformation in addition to standard random flips, random crops of $448\times448$, random brightness and random contrast augmentations.
We report results for each run using the best checkpoint model. Since the cross entropy loss does not precisely represent the IoU metric we consider both the validation IoU and loss when selecting the best model.
For all of our experiments we fine-tuned the learning rate for each setting and the results are averaged over 5 runs, unless otherwise specified.

\begin{figure}[t]
\begin{center}
\centerline{\includegraphics[width=1.0\columnwidth]{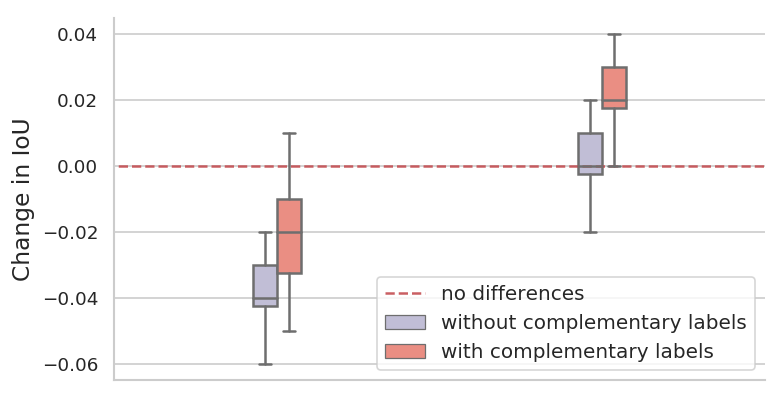}}
\vspace{-2mm}
\begin{scriptsize}
\hspace{5mm} \textsc{Expert 2} \hspace{25mm} \textsc{Expert 3} 
\end{scriptsize} \\
\vspace{-2mm}
\caption{Complementary labels provide robustness to annotator bias. The training data was annotated by \textsc{Expert 1}, biasing the models towards this expert. Test results evaluated on annotations from the other experts show that models provided with complementary labels performed consistently better (indicating robustness to bias) than those without (measured by change in IoU).}
\label{fig:ANON_generalization_gap}
\end{center}
\vspace{4mm}
\end{figure}

\subsection{Results}

\textbf{Do complementary labels help?}
Our main results appear in Figure \ref{fig:ANON_IoU_absolut}, where we quantify the effects of adding complementary labels measured by IoU.
Evidently, training with complementary labels outperforms the case where only expert tumor annotations are used by a large margin. 
Although the absolute IoU performance is relatively low, inter-expert agreement is also low ($\approx$.67), and adding complementary labels results in a 57\% relative gain in IoU towards the level of expert agreement (Table
\ref{tab:confusion}).
Segmentation visualizations appear in Figure \ref{fig:ANON_viualizations}, showing especially noticeable benefits when very little training data is available.

\textbf{What is the effect of changing the amount of training examples?}
From the results in Figure \ref{fig:ANON_IoU_absolut} it is evident that the benefit of complementary labels is magnified as we move towards lower data regimes. Models trained with expert and complementary labels from 50 patients outperform models trained on only expert labels from 130 patients.  

\textbf{Do complementary labels provide robustness against annotator bias?}
A major challenge facing intelligent diagnostic systems is disagreement between experts \cite{expert_disagreement}. Because expert annotations are costly, most medical imaging datasets are annotated by only a few experts. 
The \textsc{Csaw-S} training set was annotated solely by \textsc{Expert 1}.
This can introduce bias in the model towards the opinion of that expert.
To test if adding complementary labels introduces robustness to annotator bias in our models, we measure performance using test set annotations from the other experts.
In Figure \ref{fig:ANON_generalization_gap} we report the change in IoU when evaluating using annotations from \textsc{Expert 2} and \textsc{Expert 3} compared to \textsc{Expert 1}.
In both cases, adding complementary labels increases robustness against annotator bias.
Interestingly, \textit{both} models performed better when evaluated on \textsc{Expert 3}'s annotations.

\vspace{-1mm}
\section{Further Studies Using Public Data}
\label{experiments}

\begin{figure}[t]
\begin{center}
\centerline{\includegraphics[width=1.0\columnwidth]{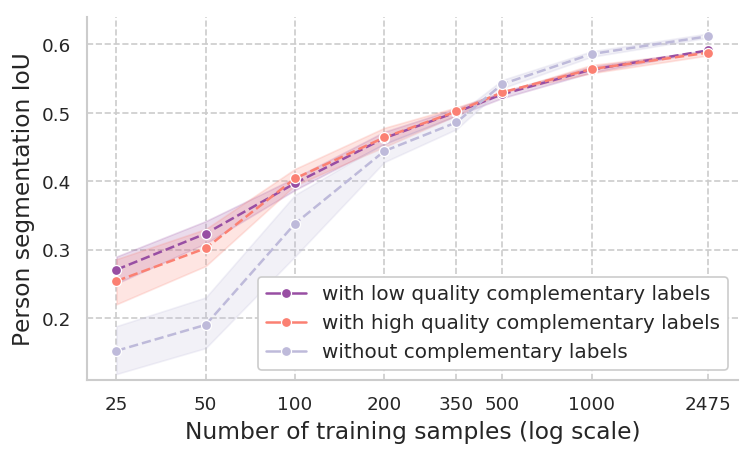}}
\vspace{-3mm}
\caption{Results on \textsc{Cityscapes} using the \textit{person} class as the expert class.
Performance with (red) and without (blue) complementary labels is measured for varying numbers of training images over five repetitions (shaded area indicates 95\% CI). An additional comparison to examine the impact of low-quality complimentary labels (violet) shows little effect.
}
\label{fig:city_coarse_and_fine_and_none}
\end{center}
\end{figure}

We confirm our findings on two publicly available datasets, \textsc{Cityscapes} and \textsc{Pascal Voc}, and also delve deeper -- using these well-known datasets to investigate several interesting properties of complementary labels.

\subsection{Public Datasets}
\label{cityscapes}

\textsc{Cityscapes} 
\cite{cityscapes_dataset}, is a sizable public dataset of urban street scenes with a large and diverse set of annotations. While the protocol for most segmentation research is to use 19 of the 34 annotated classes, in the interest of better understanding complementary labels, we use all 34 classes.

Although expert knowledge was not necessary to annotate \textsc{Cityscapes}, our goal is to test the effects of complementary labels on an expert task. We simulate this by choosing the \textit{person} class as the target class, so the goal becomes person segmentation. The other 33 classes are either treated as complementary labels, or merged into a single \textit{background} class for the baseline. We selected \textit{person} because of the high intra-class variance, relatively small size, and moderate image frequency. Additionally, the \textit{rider} class (\textit{e.g.} cyclists, motorcycle riders) presents an opportunity to examine performance when confusing objects are present. 
We randomly select 500 images from the official training set to use as the validation set, and we use the rest for training.
For testing, we use the official validation set of fine annotations.

\begin{figure}[t]
\begin{center}
\centerline{\includegraphics[width=1.0\columnwidth]{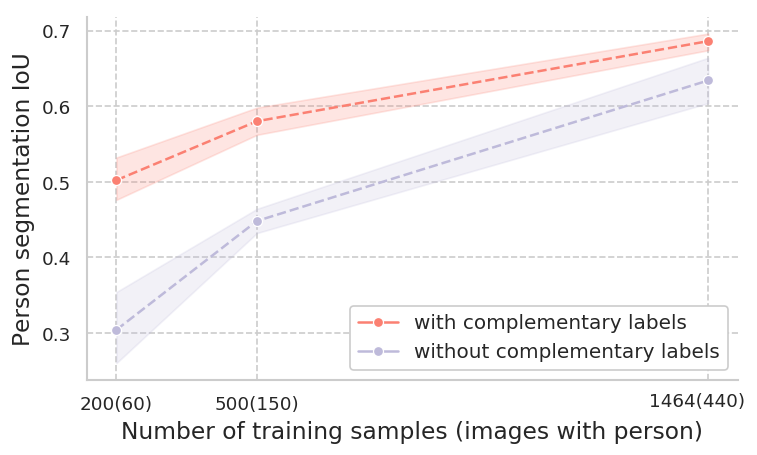}}
\vspace{-2mm}
\caption{Results on \textsc{Pascal Voc} using the \textit{person} class as the expert class.
Because \textit{person} occurs rarely, measurements are taken at different numbers of training samples than the previous experiments.}
\label{fig:voc_results}
\end{center}
\end{figure}

\textsc{Pascal Voc} \cite{voc_dataset} is a well-studied object recognition dataset composed of 21 object classes and a background class. 
Here, we also choose the \textit{person} as the main target and we consider the other 20 classes as either complementary labels or background.
In \textsc{Pascal Voc}, the \textit{person} class is present in only $30\%$ of the training images, which allows us to investigate how complementary labels perform for expert tasks on rare events. 
We used the official \texttt{train-2012} as training set and we sampled 500 images from the \texttt{test-2012} as our validation set. 
We used the remaining 949 images from \texttt{test-2012} to evaluate the final performance.

\subsection{Implementation Details}
For consistency, we kept the training procedure constant throughout our experiments and used the training settings outlined in Section \ref{applications}, with the following exceptions: random rotation and elastic augmentations were omitted, and the full images were resized to $512\times512$ instead of using patches.
We note that settings optimized for \textsc{Csaw-S} might be sub-optimal for these datasets (see Appendix), but we opted to maintain the same settings for better comparability.

\subsection{Results}

\textbf{Are the benefits of complementary labels observed in other data?}
We confirm our previous findings on medical images with data from \textsc{Cityscapes} and \textsc{Pascal Voc} in Figures \ref{fig:city_coarse_and_fine_and_none} and \ref{fig:voc_results} respectively.
We note that, although they are not directly comparable, the IoU gap between models with/without complementary labels is nearly the same for \textsc{Csaw-S} and \textsc{Cityscapes}, and the gap nearly doubles for \textsc{Pascal Voc}.
The size of these datasets allowed us to explore how the trend evolves further from the low-data regime.
Interestingly, on \textsc{Cityscapes} there appears to be a crossover when the network is provided with approximately 400 examples where complementary labels seem to hurt performance rather than help.

\begin{figure}[t]
\begin{center}
\centerline{\includegraphics[width=\columnwidth]{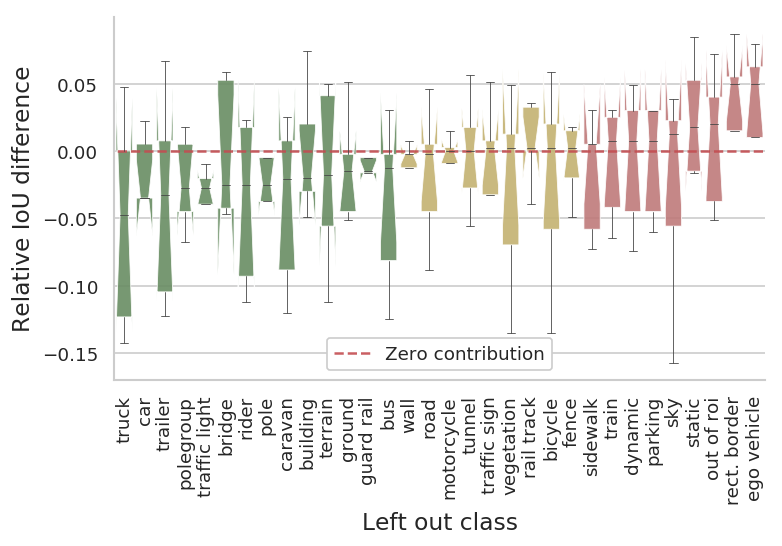}}
\vspace{-2mm}
\caption{Contributions of individual complementary labels. We conducted a leave-one-out experiment where the network was trained in turn using every complementary label in \textsc{Cityscapes} except one. The importance of each label is measured by the relative change in IoU for person classification when the label is omitted. Labels marked in green, which tend to occur frequently in the data with diverse appearances, improve performance. Labels marked in yellow have little discernible effect, and labels marked in red hurt performance. These tend to be omnipresent and trivial (\textit{e.g.}~ego vehicle and rectification border).}
\label{fig:city_label_importnce}
\end{center}
\vspace{1mm}
\end{figure}

\textbf{Do all complementary labels contribute equally?}
We investigate whether all complementary labels contribute positively, and to what degree each complementary label helps. 
To this end, we conducted a leave-one-out experiment where the network was trained with every complementary label except one, which was merged to the background. In this way, we can measure the importance of the left-out class by the change in segmentation IoU.
As seen in Figure \ref{fig:city_label_importnce}, \textit{there is a clear disparity between various complementary labels.}
Most of the complementary labels contribute positively -- removing them hurts the network's performance (green). The effect is unclear for nine (yellow), and for the rest there is a clear advantage in removing them (red).

Looking more closely at Figure \ref{fig:city_label_importnce}, we infer that the labels which seem to contribute most are those that appear frequently with diverse appearances (truck, car, traffic light). 
Also, smaller objects that could potentially be confused with the \textit{person} class appear to help (pole groups). 
The least helpful labels seem to be omnipresent and trivial (\textit{e.g.}~ego vehicle and rectification border).
Finally, we note that \textit{rider}, which can be easily confused with \textit{person}, is important for the network to avoid false positives (further experiments in Appendix).

\begin{figure}[t]
\begin{center}
\centerline{\includegraphics[width=1.0\columnwidth]{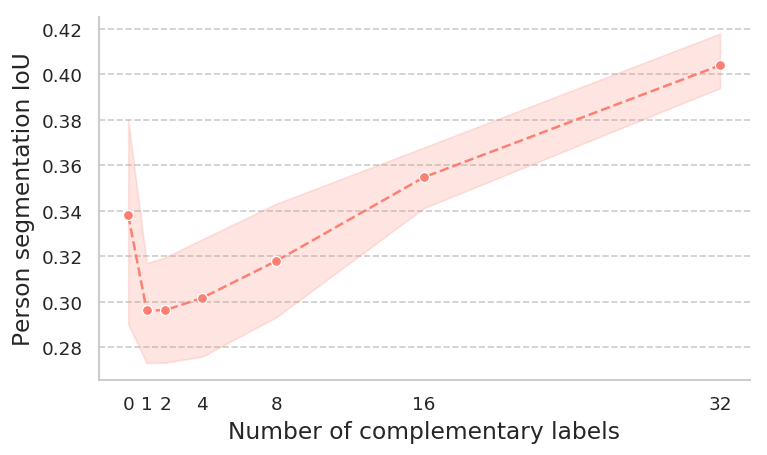}}
\vspace{-2mm}
\caption{The effect of adding new complementary labels. We measure how IoU performance changes on \textsc{Cityscapes} for a steadily increasing number of randomly selected complementary classes. Interestingly, adding only a few classes hurts performance, but once a sufficient number of labels is reached there is a clear advantage. These results, with those of Figure \ref{fig:city_label_importnce}, suggest that the choice of which labels to include is important.}
\label{fig:city_IoU_vs_labels}
\end{center}
\vspace{1mm}
\end{figure}

\textbf{What is the effect of adding new complementary labels?}
To measure the impact that the number of complementary classes has on the model's performance, we conducted a series of experiments on \textsc{Cityscapes} where the number of complementary labels provided to the network is steadily increased, $N = \{0,1,2,4,8,16,32\}$. The labels added in each run are randomly selected, and each experiment is repeated 5 times. The number of training examples was fixed to 100 and \textit{person} was again used as the expert target. We observe an interesting trend in Figure \ref{fig:city_IoU_vs_labels}, where adding only a few random complementary labels hurts performance. But once a sufficient number of labels are used, there is a clear advantage to adding them. These results, combined with those of Figure \ref{fig:city_label_importnce}, imply that care should be taken when choosing which complementary labels to include. Either that, or a sufficient number should be collected to overcome adverse effects from certain classes.

\begin{figure}[t]
\vskip 0.2in
\begin{center}
\vspace{-4mm}
\begin{tabular}{@{}c@{\hspace{0.5mm}}c@{\hspace{0.5mm}}c@{}}
    \includegraphics[width=0.333\columnwidth]{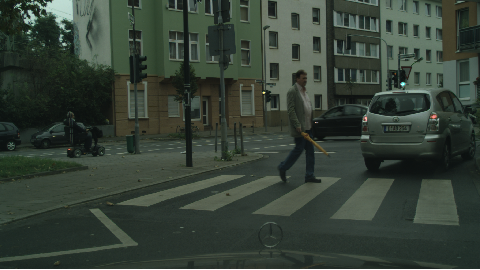} & 
    \includegraphics[width=0.333\columnwidth]{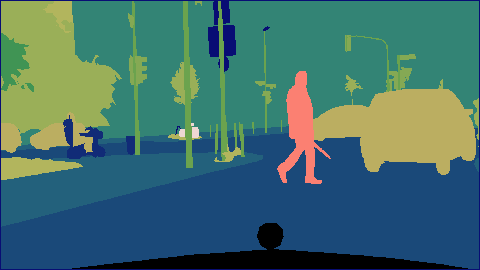} & 
    \includegraphics[width=0.333\columnwidth]{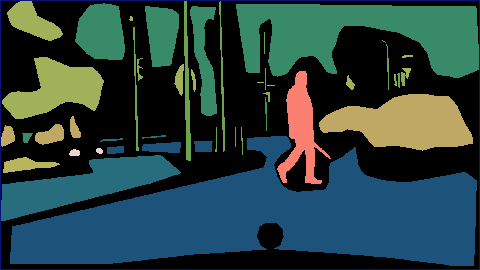}\\[1mm] 
    {\scriptsize original} & {\scriptsize high quality} & {\scriptsize low quality} \\
\end{tabular}
\vspace{-2mm}
\caption{Testing robustness to label quality. To test how complementary label quality affects performance, we compare how models trained with fine labels fare against models trained with coarse labels on \textsc{Cityscapes}. Note that high quality labels are given for the expert class, \textit{person}, in both cases. Results in Figure \ref{fig:city_coarse_and_fine_and_none} show little effect from reducing quality of complementary labels.}
\label{fig:city_annotation_example}
\end{center}
\vspace{4mm}
\end{figure}

\textbf{How does the quality of complementary labels affect performance?}
We demonstrated in Section \ref{methods} that complementary labels do not require expert knowledge, and also noted that some the labels in \textsc{Csaw-S} contain errors and noise. 
This leads us to the question: does quality of the complementary annotations affect performance?
Luckily, the \textsc{Cityscapes} dataset includes two sets of annotations for semantic segmentation, the fine and coarse set.
The fine set (high quality), seen in Figure \ref{fig:city_annotation_example}, consists of high quality dense pixel annotations whereas the coarse set (low quality) includes approximate polygonal annotations -- which omit regions of objects in many cases and in some cases actually mislabel objects.
We tested the performance of the model trained with low-quality annotations and compared it to the high-quality annotations.
To ensure the expert task was not affected, we used high-quality annotations for the \textit{person} class in both cases.
As we can see from Figure \ref{fig:city_coarse_and_fine_and_none}, the complementary labels do not need to be accurate.
Unexpectedly, for the extreme low regimes, the low-quality annotations resulted in slightly higher IoU scores. A possible explanation is that the coarse labels avoid boundary regions between objects, which may help learning.

\begin{figure}[t]
\begin{center}
\centerline{\includegraphics[width=\columnwidth]{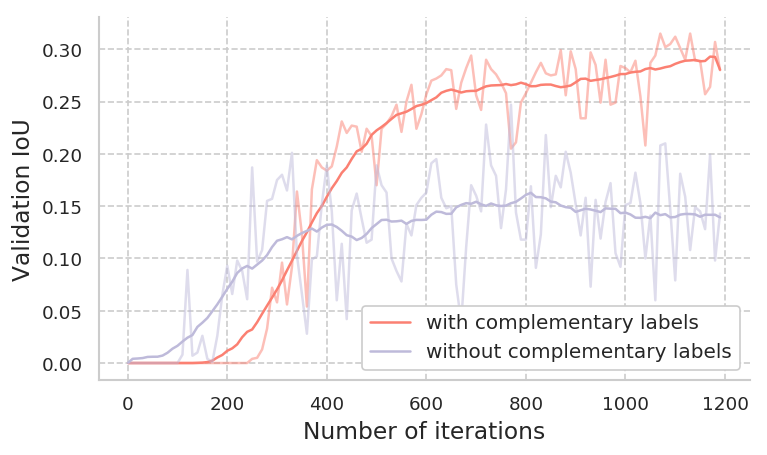}}
\vspace{-3mm}
\caption{Complementary labels improve training stability in low data regimes. As seen in this IoU evolution curve training on 25 images from \textsc{Cityscapes}, training with complementary labels yields smoother and clearer learning curves, which helps to identify signs of overfitting.}
\label{fig:IoU_evolution}
\end{center}
\end{figure}

\textbf{Do complementary labels improve training stability?}
Over the course of our experiments, we noticed that the addition of complementary labels resulted in more stable training in the low data regime. Training with complementary labels yields smoother and clearer learning curves, as seen in Figure \ref{fig:IoU_evolution}, which makes it easier to identify signs of overfitting.

\textbf{Do complementary labels provide robustness to domain shifts in the training data?}
A common issue in many datasets -- especially in medical applications -- are domain shifts within the data. 
For example, medical images can be acquired from a small number of clinics with different devices.
The \textsc{Cityscapes} images were collected from 18 different German cities and Zurich.
This phenomenon can cause generalization issues if the training data is not representative of the true distribution.

As a final investigation, we test if adding complementary labels improves model robustness to domain shifts.
We set up an experiment in which domain shifts are artificially imposed in the training data as follows. 
An ordered training set is created by shuffling images individually from each city, and then placing them in a random order grouped by city. 
For example, this may result in a training set with images from Stuttgart, then Aachen, Hamburg, etc. 
Then, we repeat the experiments for \textit{person} segmentation. The models are trained using subsets of the randomly ordered and shuffled training sets, with the same schedule as our main experiments $N=\{25,50,100,200,350,500,1000,2475\}$.
We repeat this 5 times for each $N$.
The result of this procedure is that models trained with $N=25$ or $N=50$ will only see data from a single city, but will be tested on a set containing all cities (each city contains between 77 and 259 images).  
In this way, we artificially impose a domain shift which lessens as more training data is added (more cities will appear).

Our results appear in Figure 5 and Figure 6 the Appendix. We find that adding complementary labels still improves performance in the presence of domain shifts. Although the absolute IoU performance is lower than in in Figure \ref{fig:city_coarse_and_fine_and_none}, the performance gap between models with and without complementary labels widens to some degree. Note that the curve for the model trained without complementary labels shows more variance, and is more sensitive to steps where new cities are added. 

\section{Conclusions}
\label{conclusions}

In this work we consider a semantic segmentation task and show that adding inexpensive and seemingly uninformative labels can significantly increase the model's generalisation  under low data regimes. We demonstrate the effects of these complementary labels on an expert medical task and on the \textsc{Cityscapes} and \textsc{Pascal Voc} datasets.
We release a new dataset, \textsc{Csaw-S}, along with this study which contains valuable mammography images with labels from multiple experts and non-experts that can be used to replicate our study and for other tasks in conjunction with its larger companion dataset.

We identify several interesting properties of complementary labels. First, these labels yield larger benefits when data is scarce. This property is critical in domains such as medicine where data gathering and annotation costs are often prohibitive. Separately, we find that complementary labels need not be of high quality, which suggests crowd-sourcing solutions or automation may be utilized for additional cost savings. We note that not all labels are equally useful, a fact that can help guide the annotation design process -- we witnessed that certain trivial objects hurt performance. Complementary labels seem to provide several forms of robustness to some degree: against annotator bias, against domain shift, and increased training stability. 

Complementary labels are a simple and effective means to improve performance in low-data settings. We believe they should be an essential tool in every practitioner's toolkit. However, further research is required to gain a better understanding of the mechanisms that result in this behavior.

\section*{Acknowledgements}

This work was partially supported by the Wallenberg Autonomous Systems Program (WASP), the Swedish Research Council (VR) 2017-04609, and Region Stockholm HMT 20170802. 
We would like to thank Josephine Sullivan for the fruitful discussions.
Finally, we thank the reviewers for their constructive criticism and comments.

\bibliographystyle{icml2020}
\bibliography{References}
\newpage

\twocolumn[
{\center\baselineskip 18pt
    \vskip .25in{\Large\bf
    Appendix: Adding Seemingly Uninformative Labels Helps in Low Data Regimes
}\vskip .52in}
]

\appendix
\icmltitlerunning{}
\section{The \textsc{Csaw-S} Dataset}
\label{ANON-S2}

As one of our contributions, we release the \textsc{Csaw-S}
dataset, which can be found at \href{https://github.com/ChrisMats/CSAW-S}{https://github.com/ChrisMats/CSAW-S}.
\textsc{Csaw-S} is a curated dataset containing mammography images with annotations of breast cancer and breast anatomy from experts and non-experts. It can be used to replicate our study, repurposed for other semantic segmentation tasks, or used in conjunction with other breast cancer datasets (\textit{e.g.}~DDSM \cite{DDSM_dataset}, INbreast \cite{INbreast_dataset}) to form a large repository of mammograms with tumor annotations.

\subsection{Collection}

\begin{figure}[t!]
\begin{center}
\begin{tabular}{@{}c@{\hspace{-1mm}}c@{}}
 \includegraphics[width=0.47\linewidth]{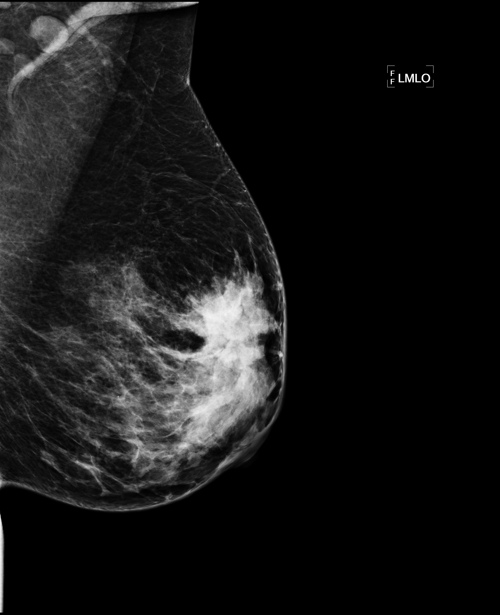}    &
 \includegraphics[width=0.47\linewidth]{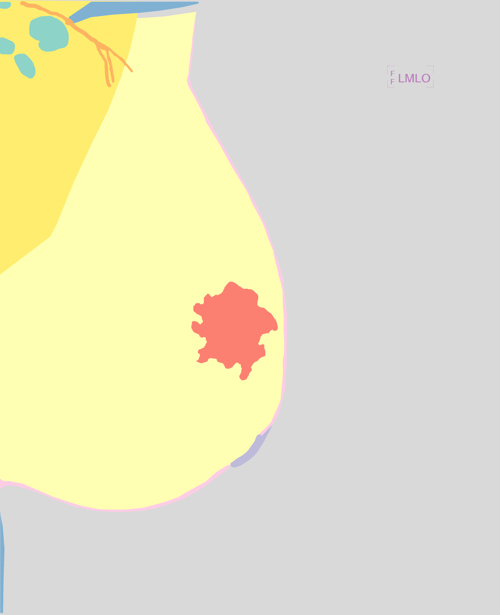}\\[-1mm]
  \includegraphics[width=0.47\linewidth]{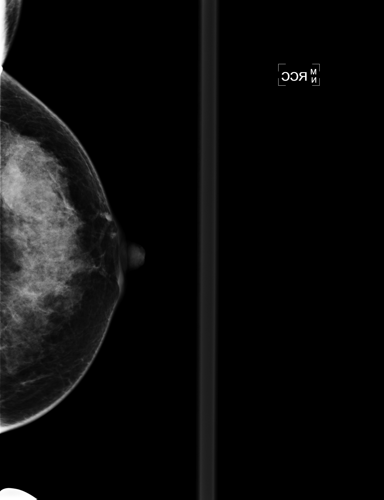}    &
 \includegraphics[width=0.47\linewidth]{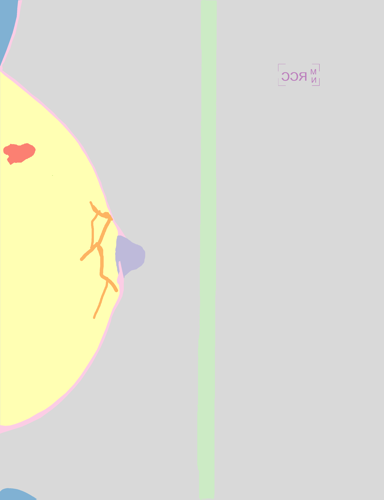}\\[-1mm]
  \includegraphics[width=0.47\linewidth]{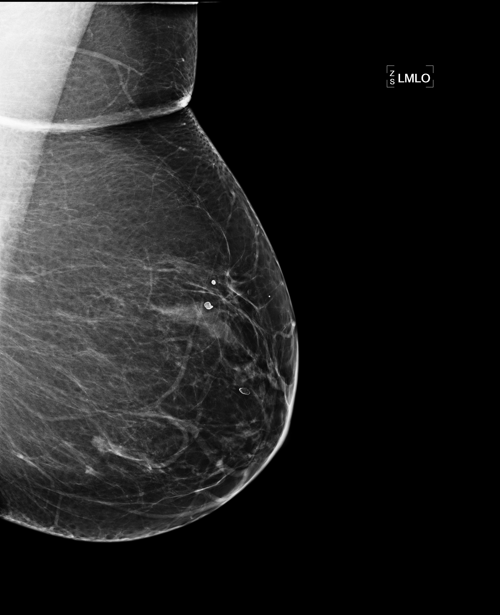}    &
 \includegraphics[width=0.47\linewidth]{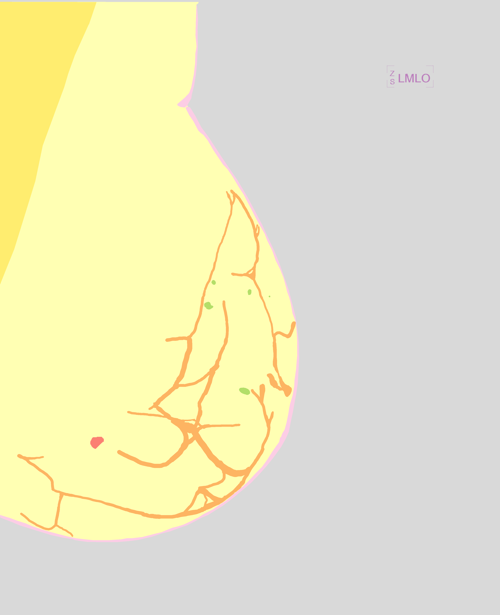}\\[2mm]
 \multicolumn{2}{l}{\hspace{-3mm} \includegraphics[width=.94\linewidth]{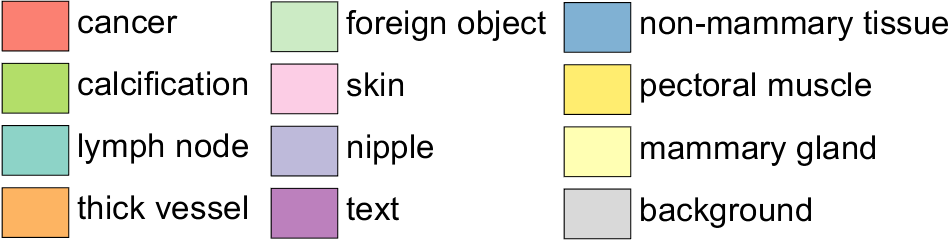}}\\
\end{tabular}
\vspace{-1mm}
\caption{Three randomly selected examples from the \textsc{Csaw-S} dataset, which contains screening mammography images from 172 cases of breast cancer. Expert radiologist labels for cancer and \textit{complementary labels} of breast anatomy made by non-experts are provided for all 342 images. The non-expert labels are in some cases imprecise or inaccurate (\textit{e.g.} skin folds in top \& bottom rows, or the improperly labeled object in lower-left of the middle row).}
\label{fig:complementary_labels2}
\end{center}
\end{figure}

The \textsc{Csaw-S} dataset contains mammography screening images from 172 different cases of breast cancer. 
It is split into a training/validation set containing 312 images from 150 patients and a test set of 26 images from 23 different patients. 
The data was split to ensure that roughly the same distribution of classes appears in the training and test splits.
In total, 338 high-resolution grayscale images of both MLO (Mediolateral-Oblique) and CC (Cranial-Caudal) views appear in the dataset. 
The screening images composing the dataset were selected from \textsc{Csaw}, a large corpus of screenings gathered from Hologic devices in Stockholm between 2008 and 2015 \cite{dembrower2019multi}. The dataset is annotated with 12 classes (see Figure \ref{fig:complementary_labels2}) including the \textit{cancer} class (the expert task), ten classes representing breast anatomy (complementary classes), and the \textit{background} class.
The cancer annotations are provided by three radiology experts. 
The choice of which classes to annotate was guided by a discussion with the radiologist experts with a semantic segmentation task in mind (labeling prominent anatomy for quality control and studying possible correlation with cancer). 
Non-experts provided annotations for the complementary classes in the training set. 
Experts provided annotations for the complementary classes in the test set but not the training set. 
The non-experts had no prior experience with mammography images, but received a short training session.
The complementary classes are highly imbalanced, with some appearing very infrequently (\textit{e.g.} lymph nodes and calcifications, see Table \ref{tab:anon_stats}).

\vspace{-1mm}
\subsection{Preprocessing}

The image files from the mammography device are in the DICOM image format. The metadata of each file contains intensity windows (center and width) that determine the proper range of displayed pixel intensities accounting for differences in acquisition such as exposure, compression, etc.
As a preprocessing step, we normalize each image using the DICOM window center and width metadata to re-scale the intensity range of the images. The re-scaling is done linearly and pixels outside the defined range are clipped.
We used this approach because, as reported in \cite{dicom_file_preproc}, the window values that are chosen by the operator or device result in consistent appearance for display. 
Many images exhibited inverted contrast (\textit{i.e.} the background was white). To rectify this, we corrected the images by inspecting the DICOM photometric interpretation attribute which determines whether the minimum pixel value is black or white. 
Finally, we convert the DICOM files to 8-bit PNG images.

\icmltitlerunning{Appendix: Adding Seemingly Uninformative Labels Helps in Low Data Regimes}

\subsection{Expert and Non-expert Annotations}

Cancer annotations for the training/validation set were provided by \textsc{Expert 1}. The complementary labels were sourced from seven non-experts with no medical training, one annotator per image. Each non-expert annotated between 15 and 109 images.

\begin{table}[t]
\vspace{-0.1in}
\caption{Class ratios in the \textsc{Csaw-S} training set by pixel count and by image frequency reveal highly imbalanced classes.}
\label{tab:anon_stats}
\vspace{1mm}
\begin{center}
\begin{small}
\begin{tabular}{@{}llll@{}}
\toprule
\textsc{Pixel count} & ($\%$) & \textsc{Image frequency} & ($\%$) \\
\midrule
background & 66.501 & background & 100.0 \\
mammary gland & 28.522 & mammary gland & 100.0 \\
pectoral muscle & 2.879 & cancer & 100.0 \\
thick vessels & 0.616 & skin & 100.0 \\
skin & 0.615 & nipple & 86.2 \\
cancer & 0.292 & text & 80.1 \\
non-mammary tissue & 0.271 & thick vessels & 77.9 \\
foreign object & 0.124 & non-mammary tissue & 55.4 \\
nipple & 0.082 & calcifications & 52.2 \\
lymph nodes & 0.063 & pectoral muscle & 50.0 \\
text & 0.031 & lymph nodes & 21.8 \\
calcifications & 0.003 & foreign object & 18.6 \\
\bottomrule
\end{tabular}
\end{small}
\end{center}
\vspace{4mm}
\end{table}

Along with each test image we provide cancer annotations from three expert radiologists: \textsc{Expert 1}, \textsc{Expert 2} and \textsc{Expert 3}. Complementary labels of the breast anatomy are provided by both experts and non-experts for the test set.  \textsc{Expert 1} and \textsc{Expert 2} provide the complementary labels (no complementary labels are provided by \textsc{Expert 3} but they may be available in the future). Complementary labels are also provided by three non-expert annotators. The test set complementary labels were not utilized in our study, but are provided for other researchers interested in semantic segmentation of medical images.

\begin{figure}[t!]
\begin{center}
\centerline{\small{Evaluation on \textsc{Expert 1} annotations}}
\centerline{\includegraphics[width=.98\columnwidth]{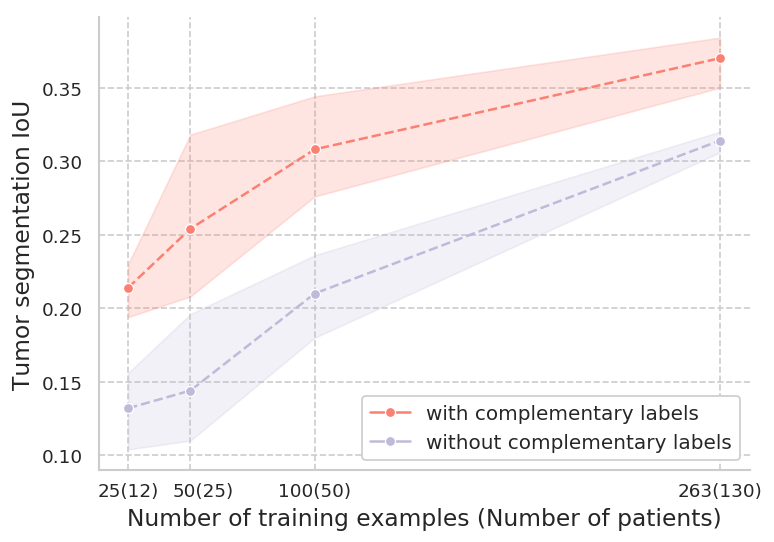}}
\vspace{1mm}
\centerline{\small{Evaluation on \textsc{Expert 2} annotations}}
\centerline{\includegraphics[width=.980\columnwidth]{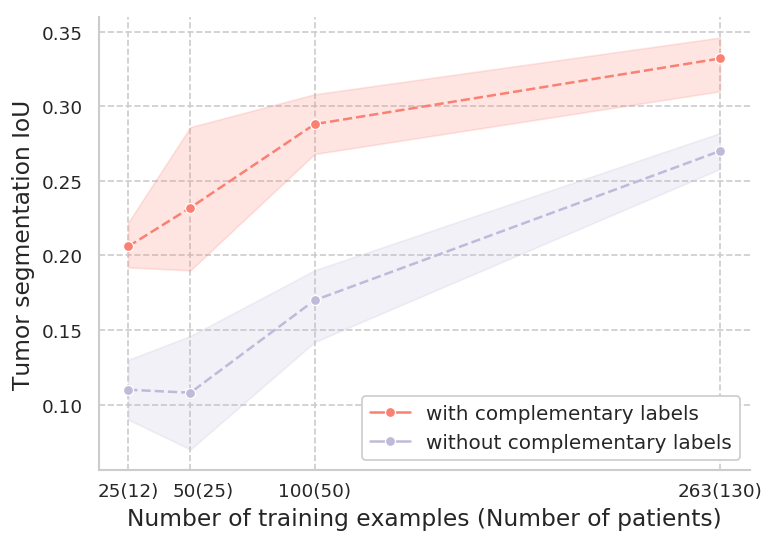}}
\centerline{\small{Evaluation on \textsc{Expert 3} annotations}}
\vspace{1mm}
\centerline{\includegraphics[width=.98\columnwidth]{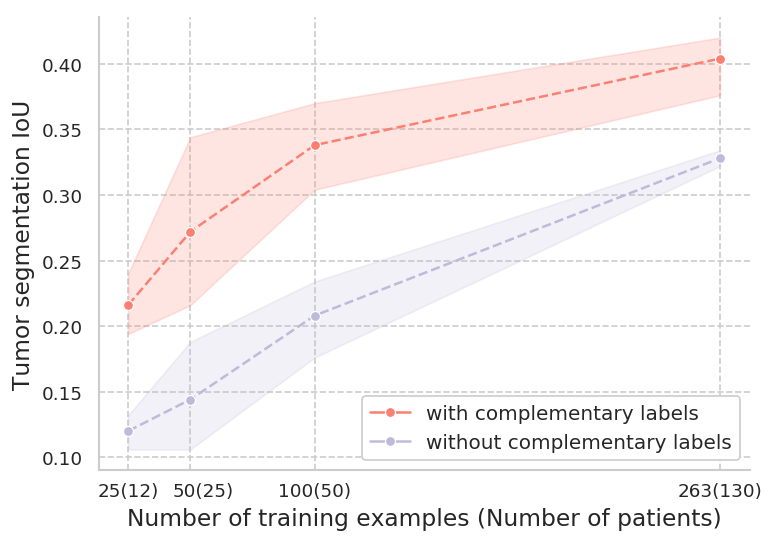}}
\vspace{-3mm}
\caption{Results on \textsc{Csaw-S} when evaluating using annotations from different experts: \textsc{Expert 1} (top) , \textsc{Expert 2} (middle) and \textsc{Expert 3} (bottom). 
The training data was annotated by \textsc{Expert 1}, biasing the models towards this expert.
Evidently, training with complementary labels results in higher IoU scores for all cases. This difference is magnified when evaluating the annotations provided by \textsc{Expert 2} and \textsc{Expert 3}, indicating an increased robustness to annotator bias when complementary labels are used. Interestingly, all models performed better when evaluated on \textsc{Expert 3}’s annotations.}
\label{fig:anon_expert_1_IoU}
\end{center}
\vspace{-2mm}
\end{figure}

Class labels of each image in the dataset are provided in the form of independent full pixel-wise binary masks for each class. These annotations were generated using QuPath \cite{bankhead2017qupath}, a tool for annotating large medical 
images. 
Annotators were instructed to completely mark each class. 
In case of overlaps, all overlapping objects were labeled. 
In this work, we do not address multi-label classification (where each pixel can be associated with multiple classes). Therefore, the 11 binary masks for each image were combined to form a single multi-class mask as follows.
Each pixel was assigned a single label by inspecting the 11 masks. Unannotated pixels were designated as background. Pixels with only one matching annotation receives the class label of the corresponding masks.
In case of overlap, the labels were determined by the following priority order (with lower values having higher priority): (1) \textit{cancer}, (2) \textit{calcification}, (3) \textit{lymph node}, (4) \textit{thick vessel}, (5) \textit{foreign object}, (6) \textit{skin}, (7) \textit{nipple}, (8) \textit{text}, (9) \textit{non-mammary tissue}, (10) \textit{pectoral muscle}, (11) \textit{mammary gland} and (12) \textit{background}. The twelve-class masks generated by this process were used to train the networks.

\begin{figure}[t!]
\begin{center}
\centerline{\includegraphics[width=1.0\columnwidth]{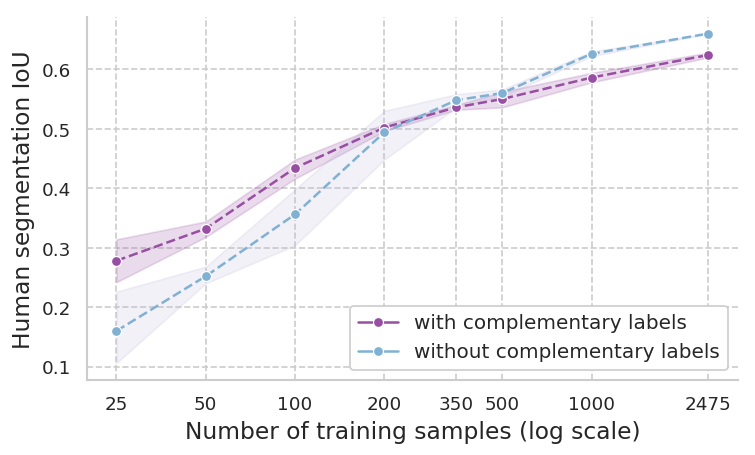}}
\vspace{-4mm}
\caption{Results after merging the confusing classes \textit{person} and \textit{rider} on the \textsc{Cityscapes} dataset into a unified label \textit{human} (the complementary classes are reduced from 32 to 31). Compared to Figure 5 in the main article, there is an absolute increase in IoU scores in both models, but the overall trend remains, and it is clear that additional complementary labels still help. }
\label{fig:human_city_IoU}
\end{center}
\end{figure}

\subsection{Class Imbalance}
The multi-class annotations for the \textsc{Csaw-S} dataset, produced as described above, are highly imbalanced (see Table \ref{tab:anon_stats}). Classes such as \textit{cancer}, \textit{skin}, \textit{mammary gland}, and \textit{background} appear in every image, but with vast differences in area (\textit{e.g.~cancer} only accounts for 0.29\% of pixels in the dataset while \textit{mammary gland} accounts for 28.3\%). Some classes are even more rare, such as \textit{lymph nodes} and \textit{calcifications} which are present in 22.2\% and 51.2\% of images (respectively), but only account for 0.064\% and 0.004\% of total pixels. These two particular classes are extremely rare in terms of image support but are also useful signs for diagnosing cancer. Therefore, these rare but important classes present an interesting segmentation challenge.

\vspace{-0.65\baselineskip}
\subsection{Generating Training Examples}

Throughout our experiments on the \textsc{Csaw-S} dataset, we generate training samples by randomly extracting $512\times512$ patches from the full resolution mammography images.
This is because (1) the high resolution of the mammograms cause memory issues during the training procedure, and (2)
the small training set sizes necessitate the use of heavy augmentations, which also causes memory issues.
To ensure more balanced class representation in the training data, we generate training examples for every image by sampling 10 center-cropped patches from locations belonging to each of the 12 classes (uniform sampling).
We perform this data generation strategy  offline once for each of the 5 experimental repetitions since it is an expensive procedure, in terms of memory and computation. The with- and without-complementary label models share the same training example crops for each experimental repetition.

During training we employ an extensive set of augmentations including rotations and elastic transformation in addition to standard random flips, random crops of $448\times448$, random brightness and random contrast augmentations on the  $512\times512$ patches.

\begin{figure}[t!]
\begin{center}
\centerline{\includegraphics[width=1.0\columnwidth]{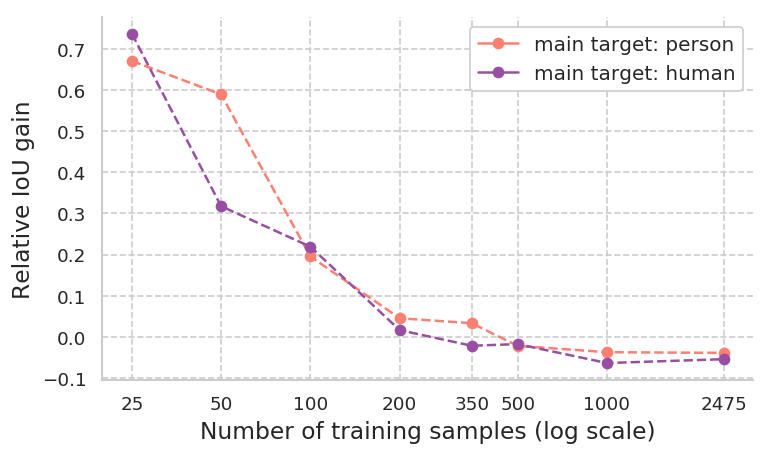}}
\vspace{-4mm}
\caption{Relative IoU gains between the with- and without-complementary label models on \textsc{Cityscapes} when the main target is \textit{person} (orange) and \textit{human} (purple). The \textit{human} class contains both \textit{person} and the confusing label, \textit{rider}.}
\label{fig:human_city_IoU_gain}
\end{center}
\end{figure}

\subsection{Evaluating Against Different Experts}

The \textsc{Csaw-S} training set contains cancer annotations only from \textsc{Expert 1}. This biases networks trained on this data towards the opinion of this expert.
The test set includes tumor annotations from \textsc{Expert 1} and two additional expert radiologists. As seen in the main article, agreement between the experts is relatively low ($\approx$0.67), therefore we expect the IoU scores to differ when evaluating on annotations from the other experts.
As we can see in Figure \ref{fig:anon_expert_1_IoU}, the complementary labels result in increased performance over the baseline, regardless of which expert is considered.
Furthermore, we can see that the generalization gap between the models trained with and without complementary labels increases as we evaluate on \textsc{Expert 2} and \textsc{Expert 3}. This indicates that the model with complementary labels is more robust to annotator bias.
Interestingly, all models performed better when evaluated on \textsc{expert 3}’s annotations.

\begin{figure}[t!]
\begin{center}
\centerline{\includegraphics[width=1.0\columnwidth]{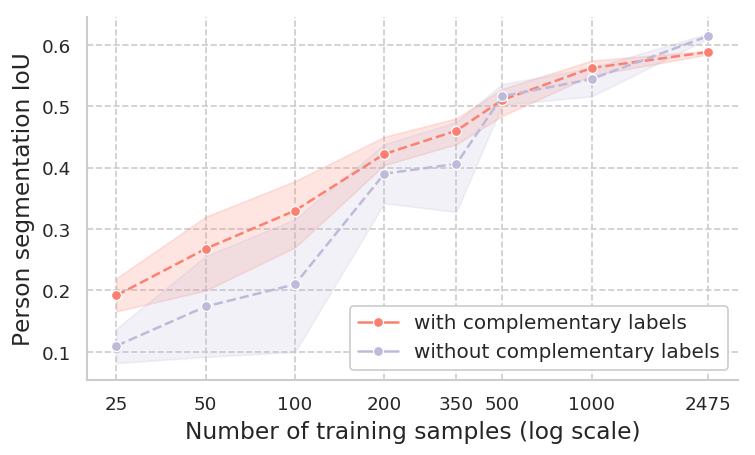}}
\vspace{-2mm}
\caption{Results when imposing a domain shift between the training and test data by ordering the training examples. Compared with Figure 5 from the main text, we note that the domain shift causes an absolute drop in performance for both models, but the performance gap remains steady.}
\label{fig:city_unordered_iou}
\end{center}
\end{figure}

\section{Sanity Checks For the Training Procedure}
\label{other_things_we_tried}
We ran a series of sanity checks to satisfy any concerns regarding the standard training protocol used in our experiments. We report that our finding -- a significant generalization gap between models with and without complementary labels -- remains consistent regardless of the training technique used.

In detail, we tested the following variations to our standard training procedure:
\begin{itemize}
    \vspace{-3mm}
    \item We replaced the DeepLabv3 segmentation model with FCN-32 \cite{fcn}. We found no change in the generalization gap.
    \vspace{-2mm}
    \item We replaced GroupNorm with BatchNorm. We found no change in the generalization gap.
    \vspace{-2mm}
    \item Instead of using \textsc{ImageNet} pretrained models, we randomly initialized the weights. We found no change in the generalization gap, although there was a significant drop in performance for both cases.
    \vspace{-2mm}
    \item We froze the normalization statistics \textit{1)} throughout the training process and \textit{2)} for the first 60\% of the training iterations and fine tune for the rest 40\%. We found that freezing the normalization statistics did not help the for \textsc{Csaw-S} dataset but for the \textsc{Cityscapes} and \textsc{Pascal Voc} there were significantly large improvements, especially towards the extreme low data regime. As has been reported before \cite{transfusion, rethinking_imagenet_pretrain}, this is because the statistics dramatically change as we move from the natural domain to the medical one.
    \vspace{-2mm}
    \item We extended the default augmentations with random scale and resizing as well as elastic transforms. 
    As expected, augmentations resulted in increased generalization for every case. Nonetheless, we found that heavy augmentations favour the models trained with complementary labels slightly more.
    \vspace{-2mm}
    \item We evaluated the importance of complementary labels when the main target is trivial (e.g. class \textit{sky} and \textit{ego vehicle} in \textsc{Cityscapes}).
    We consider a class as trivial when it has many of the following traits which make it easily distinguishable: little texture variation, regular shape, clear edges, and large pixel-count per image.
    For these cases, we find high IoU scores even when limited examples are present.
    We also found that the most important factor is the normalization method, and gains from the complementary labels only persist in the extreme low data regime (less than 10-20 training examples).
    \vspace{-2mm}
\end{itemize}

Finally, we note that we tuned our training settings for the case where only expert annotations are available (no complementary labels) and set them as default to ensure that the gains from the complementary labels are valid.
We did not further tune the settings for models with complementary labels, so this likely leads to sub-optimal settings and an under-reporting of the actual generalization gap.

\begin{figure}[t!]
\begin{center}
\centerline{\includegraphics[width=1.0\columnwidth]{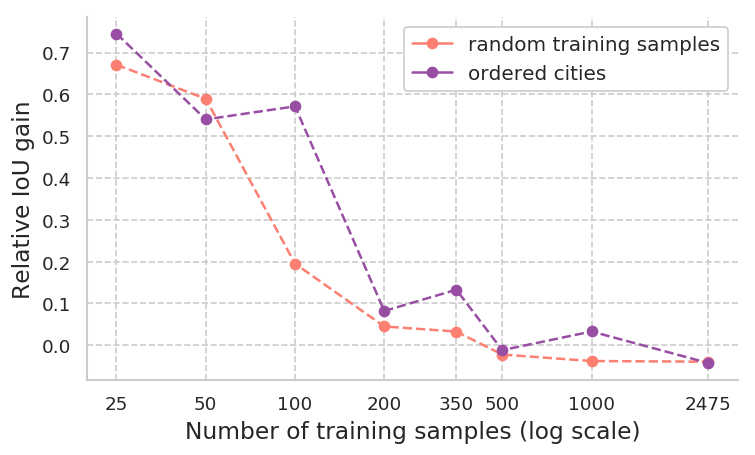}}
\vspace{-2mm}
\caption{Relative IoU gains (between models trained with and without complementary labels) when imposing a domain shift between the training and test data (purple), and when there is no domain shift (orange). Both cases show models trained with complementary labels. Note that, although the purple curve is noisier, there seems to be little effect on IoU performance gap between the two setups, indicating robustness to domain shifts.}
\label{fig:city_unordered_gains}
\end{center}
\end{figure}

\section{Easily Confused Classes}

The class \textit{person} is easily confused with the class \textit{rider} in the \textsc{Cityscapes} dataset.
We investigated whether gains from the complementary labels for the \textit{person} segmentation task are due to the explicit modeling of the \textit{rider} class.
In other words, are the observed effects attributed to complementary labels actually due to the modeling of easily confused classes? Intuitively, explicitly modeling the \textit{rider} class, which would otherwise be part of the background (in the without-complementary labels case), should help reduce false positives.
To investigate, we merged the annotations of the classes \textit{person} and \textit{rider} into a unified label \textit{human} and repeated the experiments.
Although the absolute IoU scores for both models improved (Figure \ref{fig:human_city_IoU}), the relative IoU gains from the complementary labels showed no appreciable change (Figure \ref{fig:human_city_IoU_gain}).
Thus, we infer that although the explicit modeling of confusing classes is helpful (see label importance in the main text), it is not the only reason that including complementary labels are beneficial in low data regimes.

\begin{table}[t!]
\caption{Class ratios in the \textsc{Pascal Voc} dataset by pixel count and by image frequency.}
\label{tab:voc_stats}
\vskip 0.15in
\begin{center}
\begin{small}
\begin{tabular}{llll}
\toprule
\textsc{Pixel count}  & ($\%$) & \textsc{Image count} & ($\%$) \\
\midrule
background  &  64.7 &  ambiguous  &  99.9 \\
ambiguous  &  5.9 &  background  &  99.5 \\
person  &  5.2 &  person  &  29.9 \\
cat  &  3.1 &  chair  &  9.9 \\
bus  &  2.0 &  cat  &  8.9 \\
dog  &  1.9 &  car  &  8.5 \\
train  &  1.8 &  dog  &  8.3 \\
car  &  1.6 &  bird  &  7.2 \\
sofa  &  1.5 &  sofa  &  6.4 \\
motorbike  &  1.3 &  aeroplane  &  6.0 \\
dining table  &  1.3 &  bottle  &  5.7 \\
chair  &  1.2 &  train  &  5.7 \\
horse  &  1.1 &  tv/monitor  &  5.7 \\
bird  &  1.0 &  dining table  &  5.6 \\
tv/monitor  &  1.0 &  motorbike  &  5.5 \\
sheep  &  1.0 &  potted-plant  &  5.5 \\
cow  &  0.9 &  boat  &  5.3 \\
aeroplane  &  0.9 &  bus  &  5.3 \\
boat  &  0.7 &  horse  &  4.6 \\
potted-plant  &  0.7 &  bicycle  &  4.4 \\
bottle  &  0.7 &  cow  &  4.4 \\
bicycle  &  0.3 &  sheep  &  4.3 \\
\bottomrule
\end{tabular}
\end{small}
\end{center}
\vspace{4mm}
\end{table}

\begin{table}[t!]
\caption{Class ratios in the \textsc{Cityscapes} dataset by pixel count and by image frequency.}
\label{tab:cityscape_stats}
\vskip 0.15in
\begin{center}
\begin{small}
\begin{tabular}{llll}
\toprule
\textsc{Pixel count}  & ($\%$) & \textsc{Image count} & ($\%$) \\
\midrule
road  &  36.61 &  ego vehicle  &  100.0 \\
building  &  20.06 &  pole  &  98.8 \\
vegetation  &  14.35 &  static  &  98.7 \\
car  &  6.69 &  road  &  98.6 \\
sidewalk  &  5.41 &  building  &  98.5 \\
sky  &  3.04 &  vegetation  &  97.0 \\
ego vehicle  &  2.43 &  car  &  94.8 \\
static  &  1.42 &  traffic sign  &  94.4 \\
ground  &  1.29 &  sidewalk  &  94.0 \\
person  &  1.26 &  sky  &  86.3 \\
pole  &  1.18 &  person  &  78.5 \\
terrain  &  1.05 &  traffic light  &  55.8 \\
fence  &  0.77 &  terrain  &  54.9 \\
parking  &  0.63 &  bicycle  &  53.7 \\
wall  &  0.58 &  dynamic  &  44.7 \\
traffic sign  &  0.52 &  fence  &  42.1 \\
bicycle  &  0.4 &  ground  &  34.7 \\
bridge  &  0.3 &  rider  &  34.2 \\
dynamic  &  0.29 &  wall  &  31.6 \\
train  &  0.27 &  parking  &  24.4 \\
truck  &  0.23 & rect. border  &  22.2 \\
bus  &  0.22 &  unlabeled  &  19.7 \\
rect. border  &  0.22 &  motorcycle  &  17.1 \\
traffic light  &  0.19 &  truck  &  11.9 \\
rail track  &  0.19 &  bus  &  8.6 \\
rider  &  0.14 &  out of roi  &  8.6 \\
motorcycle  &  0.1 &  bridge  &  7.5 \\
tunnel   &  0.05 &  polegroup  &  7.2 \\
caravan  &  0.04 &  train  &  4.4 \\
out of roi  &  0.03 &  rail track  &  3.5 \\
trailer  &  0.02 &  trailer  &  2.5 \\
unlabeled  &  0.01 &  caravan  &  1.9 \\
guard rail  &  0.01 &  tunnel  &  0.8 \\
polegroup  &  0.01 &  guard rail  &  0.6 \\
\bottomrule
\end{tabular}
\end{small}
\end{center}
\vspace{4mm}
\end{table}

\section{Robustness to Domain Shifts}

As described in the main article at the end of Section 5.3, we tested if adding complementary labels improves model robustness to domain shifts. Our goal in this experiment is to test if complementary labels help when there is a domain shift between the training distribution and the test distribution. For example, if medical images acquired from certain devices appear in the training set, and images from a different set of devices appear in the test set.

We set up an experiment in which domain shifts were artificially imposed in the training data as follows. An ordered training set is created by shuffling images individually from each city, and then placing them in a random order grouped by city. 
For example, this may result in a training set with images from Stuttgart, then Aachen, Hamburg, etc. 
Then, we repeat the experiments for \textit{person} segmentation. The models are trained using subsets of the randomly ordered and shuffled training sets, with the same schedule as our main experiments $N=\{25,50,100,200,350,500,1000,2475\}$.
We repeat this 5 times for each $N$.
The result of this procedure is that models trained with $N=25$ or $N=50$ will only see data from a single city, but will be tested on a set containing all cities (each city contains between 77 and 259 images).  This represents our imposed domain shift. As more data is added, more cities will appear in the training set.

We find that adding complementary labels improve performance in the presence of domain shifts (Figure \ref{fig:city_unordered_iou}). 
Although the absolute IoU performance is lower when the domain shift is imposed than than when there is no domain shift, the performance gap between models with and without complementary labels holds and widens to some extent (Figure \ref{fig:city_unordered_gains}). 
Furthermore, while the curve for the model trained with complementary labels retains the same shape as in Figure 5 from the main text, the curve for the model trained without complementary labels shows more variance, and is sensitive to steps where new cities are added (Figure \ref{fig:city_unordered_iou}).

\begin{figure}[t!]
\begin{center}
\includegraphics[width=\columnwidth]{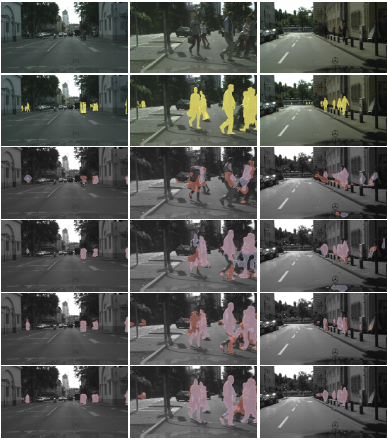} \\
\includegraphics[width=\columnwidth]{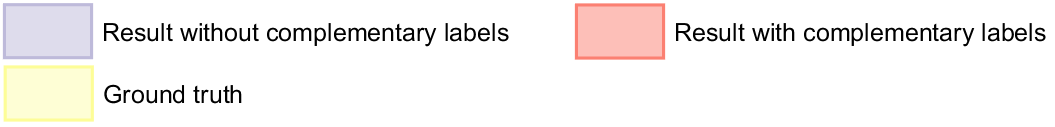}
\vspace{-7mm}
\caption{Segmentation results on \textsc{Cityscapes} show complementary labels improve performance in low-data settings. From top to bottom: the full image, fine annotations and predictions from networks trained with only the \textit{person} labels (\textit{blue}) and predictions from a network trained with both \textit{person} labels and complementary labels (\textit{red}) using $N=\{25,100,350,2475\}$ training examples. In the last row, complementary labels begin to hurt performance (the crossover in Figure 5 in the main text).}
\label{fig:City_viualizations}
\end{center}
\vspace{2mm}
\end{figure}

\section{Label Frequency on \textsc{Cityscapes} and \textsc{Pascal Voc}}

In Table \ref{tab:voc_stats} and Table \ref{tab:cityscape_stats}, we report the number of pixels and the number of images that contain each class for \textsc{Cityscapes} and \textsc{Pascal Voc}.
Note that the \textsc{Cityscapes} dataset has severe class imbalance whereas \textsc{Pascal Voc} is more balanced -- the most and least frequent classes differ by at most one order of magnitude at most (excluding the background and ambiguous class).
The greater pixel count imbalance in \textsc{Cityscapes} corroborates our qualitative observation that objects in \textsc{Cityscapes} are more likely to appear at different scales than in \textsc{Pascal Voc}. For complex classes like \textit{person}, this can make the segmentation problem significantly harder and may partially explain the performance difference between the two datasets.

\begin{figure}[t!]
\begin{center}
\includegraphics[width=\columnwidth]{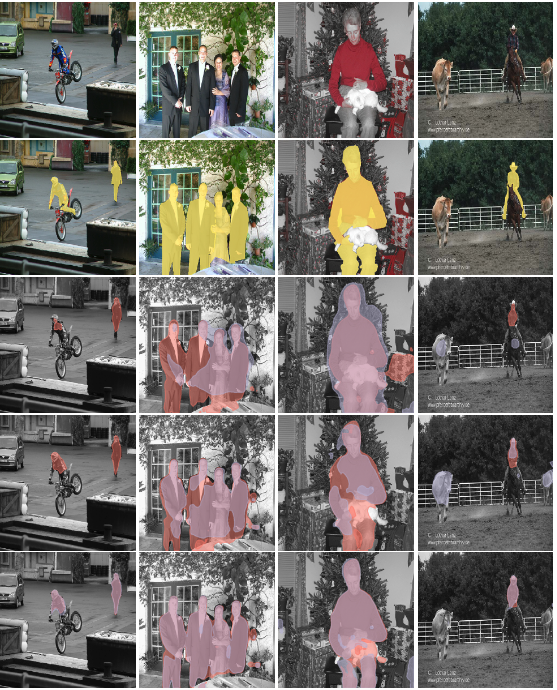} \\
\includegraphics[width=\columnwidth]{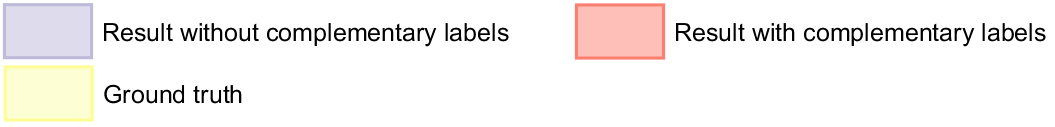}
\vspace{-6mm}
\caption{Segmentation results on \textsc{Pascal Voc} show complementary labels improve performance in low-data settings. From top to bottom: the full image, annotations and predictions from networks trained with only the \textit{person} labels (\textit{blue}) and predictions from a network trained with both \textit{person} labels and complementary labels (\textit{red}) using $N=\{200,500,1464\}$ training examples.} 
\label{fig:VOC_viualizations}
\end{center}
\vspace{3mm}
\end{figure}

\section{Segmentation Results on \textsc{Cityscapes} and \textsc{Pascal Voc}}

In Figure 5 and Figure 6 of the main text, we showed quantitatively that adding complementary labels results in increased IoU scores for \textsc{Cityscapes} and \textsc{Pascal Voc} in low data regimes.
In Figure \ref{fig:City_viualizations}, we visualize results for \textsc{Cityscapes} and confirm that  segmentations from models trained with complementary labels result in more accurate segmentation masks in low data regimes where segmentation are generally poor.
As we increase the number of training examples, we see diminishing returns when adding complementary labels. When a large number of examples are included in the training procedure (last row of Figure \ref{fig:City_viualizations}), complementary labels begin to hurt performance.

In Figure \ref{fig:VOC_viualizations}, we visualize the model predictions on \textsc{Pascal Voc} and confirm that the segmentation results also reflect the trend in IoU we reported in the main text: when limited samples are present in the training set, the addition of complementary labels results in better segmentations.


\end{document}